\pdfoutput=1

\documentclass[11pt]{article}

\usepackage[]{ACL2023}
\usepackage{amsmath}
\usepackage{amssymb}

\usepackage{times}
\usepackage{latexsym}

\usepackage[T1]{fontenc}

\usepackage[utf8]{inputenc}

\usepackage{graphicx}
\usepackage{float}
\usepackage{multirow}
\usepackage{verbatim}
\usepackage{lipsum}
\usepackage{amssymb}
\usepackage{amsmath}
\usepackage{booktabs}
\usepackage{url} 
\usepackage{enumitem}
\usepackage{mathtools}
\usepackage{subcaption}
\usepackage{fontawesome} 
\usepackage{contour}
\usepackage{textcomp}
\usepackage{amsmath}
\usepackage{microtype}
\usepackage{xspace}
\usepackage{dsfont}
\usepackage{hieroglf}

\usepackage{microtype}

\newcommand{\task}{\textsc{G4C}\xspace}
\newcommand{\gandalfEmoji}{\includegraphics[height=.9em,trim=0 .4em 0 0]{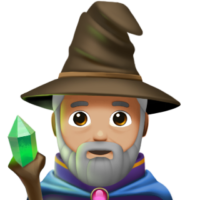}}
\newcommand{\taskWithEmoji}{\gandalfEmoji\xspace\textsc{G4C}\xspace}

\newcommand{\tasklong}{\textbf{G}enerating \textbf{G}uidance in \textbf{G}oal-Driven and \textbf{G}rounded \textbf{C}ommunication}

\newcommand{\data}{\textsc{G-Dragon}\xspace}
\newcommand{\dragonEmoji}{\includegraphics[height=.9em,trim=0 .4em 0 0]{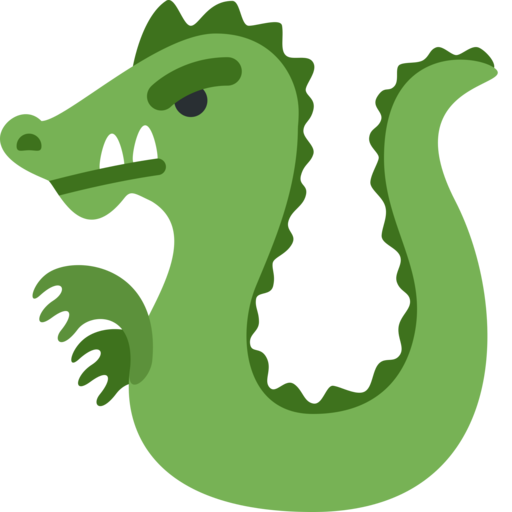}}
\newcommand{\dataWithEmoji}{\dragonEmoji\xspace\textsc{G-Dragon}\xspace}

\title{\emph{I Cast Detect Thoughts}: Learning to Converse and Guide with Intents and Theory-of-Mind in Dungeons and Dragons}

\author{
Pei Zhou$^{\heartsuit \spadesuit}$ \quad
Andrew  Zhu$^{\clubsuit}$ \quad
Jennifer Hu $^{\dagger}$ \quad
Jay Pujara$^{\spadesuit}$ \quad \\
\textbf{Xiang Ren}$^{\heartsuit \spadesuit}$ \quad
\textbf{Chris Callison-Burch}$^{\clubsuit \heartsuit}$ \quad
\textbf{Yejin Choi}$^{\diamondsuit\heartsuit}$ \quad
\textbf{Prithviraj Ammanabrolu}$^{\heartsuit\diamondsuit}$ \quad
\\
\small{$\heartsuit$ Allen Institute for Artificial Intelligence} \quad
\small{$\spadesuit$ University of Southern California}  \\
\small{$\diamondsuit$ University of Washington} \quad
\small{$\clubsuit$ University of Pennsylvania} \quad 
\small{$\dagger$ MIT}
\\{\texttt{\href{mailto:peiz@usc.edu}{peiz@usc.edu} \href{mailto:raja@allenai.org}{raja@allenai.org}}}
}
\date{}

\begin{document}
\maketitle

\begin{abstract}



We propose a novel task, \taskWithEmoji, to study \emph{teacher-student} natural language interactions in a \emph{goal-driven} and \emph{grounded} environment.
Dungeons and Dragons (D\&D), a role-playing game, provides an ideal setting to investigate such interactions.
Here, the Dungeon Master (DM), \emph{i.e.}, the teacher, guides the actions of several players---students, each with their own personas and abilities---to achieve shared goals grounded in a fantasy world.
Our approach is to decompose and model these interactions into (1) the DM's \emph{intent} to guide players towards a given goal; (2) the DM's \emph{guidance} utterance to the players expressing this intent; and (3) a \emph{theory-of-mind} (ToM) model that anticipates the players' reaction to the guidance one turn into the future.
We develop a novel reinforcement learning (RL) method for training a DM that generates guidance for players by rewarding utterances where the intent matches the ToM-anticipated player actions.
Human and automated evaluations show that a DM trained to explicitly model intents and incorporate ToM of the players using RL generates better-quality guidance that is 3x more likely to fulfill the DM's intent than a vanilla natural language generation (NLG) approach.

\end{abstract}

\section{Introduction}\label{intro}


Humans communicate with a \emph{goal} in mind and use language to reach the goal by interacting with their communication partners \emph{grounded} in a shared environment~\cite{grice1975logic,allwood1976linguistic,clark1989contributing,clark1991grounding}.
To make sure the goal is reached, we often anticipate how the partners will respond in advance to steer the conversations in the desired direction.
This ability to reason about the mental states of conversation partners -- \emph{i.e.}, theory-of-mind \citep[ToM;][]{premack1978does} -- is key to smooth and efficient communication~\cite{perner1989exploration,happe1993communicative}. 
Most existing dialogue agents, while able to produce human-like responses, often do not model communicative intents or ToM \emph{explicitly}.
In this paper, we investigate if models benefit from explicitly incorporating intents and ToM in NLG.


\begin{figure}[tb]
	\centering
	\includegraphics[width=\columnwidth]{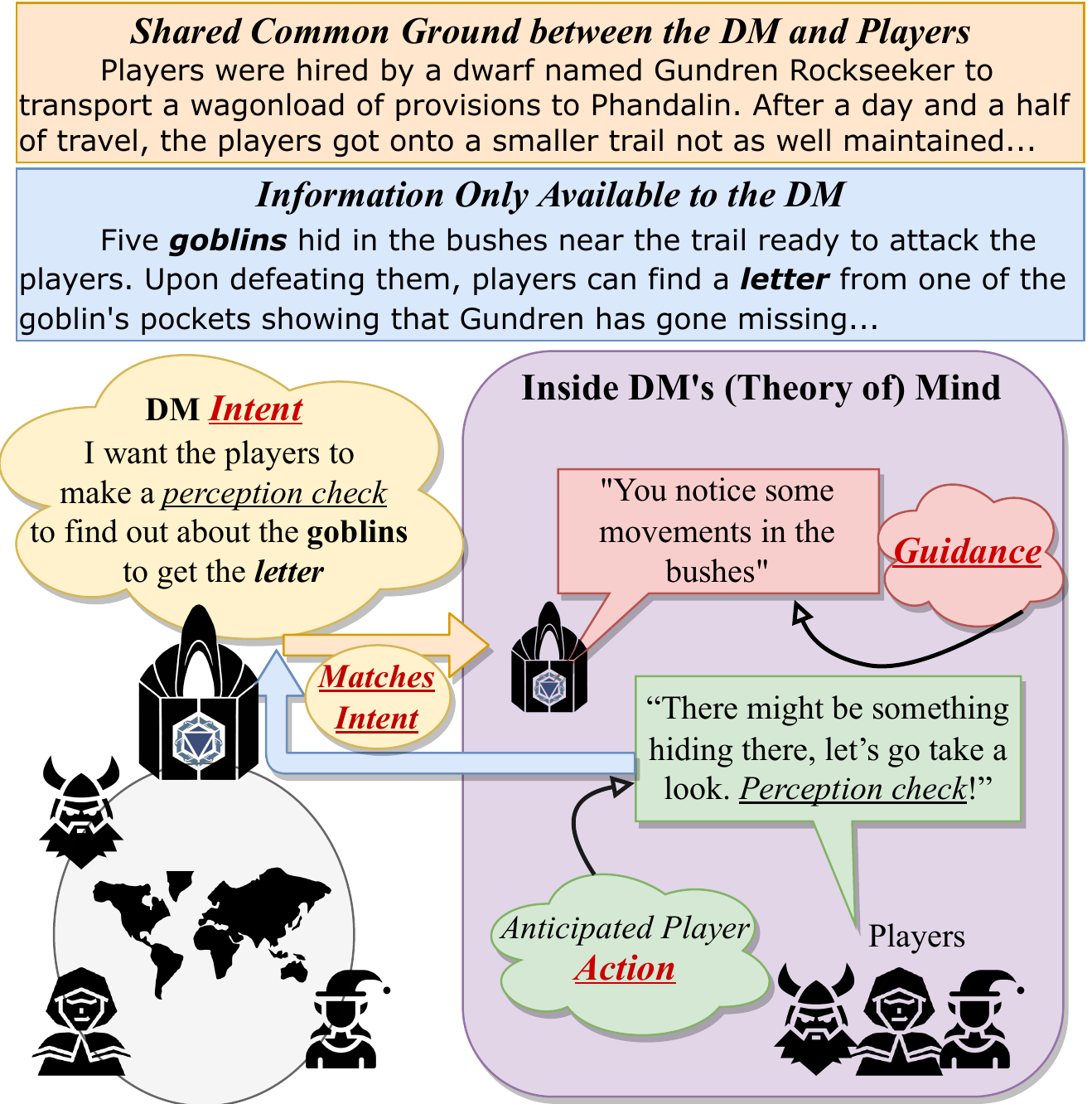}
	\caption{
    \small
	\textbf{A motivating example.} The (human) Dungeon Master (DM), knowing the desired story path, intends the players to perform actions to find out about the goblins---the first plot point that will eventually lead the players to treasure. They generate the guidance ``\emph{You notice some movements in the bushes}'' using theory-of-mind by inferring that the players will perform the desired actions upon hearing their words.}
	  	 \vspace{-0.5cm}
	\label{fig:motivation}
\end{figure}

To bridge the gap between human communication and existing dialogue models, we propose a new task \taskWithEmoji: \tasklong. 
\task considers three building blocks: \emph{intent}, \emph{guidance}, and \emph{action}. 
The task envisions a teacher with intent for specific student action, guidance uttered by the teacher, and action undertaken by the student based on the guidance and common ground.
\task evaluates the ability of a teacher to provide intentional \emph{guidance} that results in intended student actions.\footnote{Here we use \emph{actions} to indicate any linguistic behavior with intention~\cite{allwood1976linguistic}.} 
The success of the teacher's guidance depends on whether the student's subsequent action matches the teacher's \emph{intended} action. 
Using this task formulation, we analyze if the teacher has fulfilled their communicative intents explicitly by examining what the student says afterward.
\task further requires the dialogue to be grounded, meaning that both the teacher and the student are communicating with a shared environment and background.

To train models to perform \task, we use Dungeons and Dragons (D\&D) as our environment, a game that heavily relies on communication that is inherently goal-driven and grounded. 
D\&D is a role-playing game consisting of multiple player characters and a Dungeon Master (DM) who collaborate to achieve a set of goals beneficial to the players.
The DM, the narrator and host of the game, has an innate motivation to guide the players to perform a series of actions that roughly follow a pre-devised storyline culminating in a global goal, all grounded in a shared fantasy world.
An example of each component of \task in the D\&D environment (\emph{intent}, \emph{guidance}, and \emph{action}) is shown in Figure~\ref{fig:motivation}.

We construct 47k D\&D dialogues from transcripts collected by~\citet{callison-burch-tomar-et-al-2022}.
Motivated by the critical roles \emph{intents} and \emph{theory-of-mind (ToM)} play in human communication, we study the following central research question: ``\emph{Does incorporating \textbf{intent} and \textbf{ToM} make computational models better communicators?}``
Accordingly, we explore different methods for modeling \textbf{intent} and \textbf{ToM} for \task in Section~\ref{modeling}.
Specifically, we make the intents of the teacher (DM) explicit by mining intents from large language models (LLM) and appending them as additional context to guide generation.
We further propose a method to train a DM to generate guidance for a player with RL inspired by ToM. 
The DM first predicts in advance what action the player will take in reaction to the guidance and then uses this prediction as a feedback reward function to check whether the predicted action matches DM intent.

\task focuses on mimicking human communication that is goal-driven and coherent to a grounded narrative, which current automated dialogue metrics do not capture well.
As such, we further propose novel human and automated evaluation metrics to measure whether the output fits in the grounded context and fulfills communicative goals.
Our experiments show that DMs trained with explicit intents and ToM to predict how their players will react to their utterances ahead of time \emph{triples} the number of responses generated that are both grounded and fulfill the communicative intent. 
\section{\taskWithEmoji and \dataWithEmoji}\label{concept}

Here we discuss how we construct the environment for the proposed \taskWithEmoji task using a dataset of dialogues from Dungeons and Dragons (D\&D) called \dataWithEmoji. 
We start with formulating the \task task, then introduce the D\&D data, and finally present our procedure of constructing the environment using large-scale data.

\begin{table}[tb]
\centering
\resizebox{\columnwidth}{!}{
\begin{tabular}{c|l}
\textbf{Character} & \multicolumn{1}{c}{\textbf{Game Dialogue}}                                                                                                      \\ \hline
DM                 & \begin{tabular}[c]{@{}l@{}}A dwarf named Gundren Rockseeker has hired\\ you to transport a wagonload of provisions to\\ the rough-and-tumble settlement of Phandalin...\\
\textbf{You all notice some movements in the bushes} \\ \textbf{nearby the road...} \\
\end{tabular} 
\\ \hline 
Clint              & \begin{tabular}[c]{@{}l@{}}"There might be something hiding there, let’s go \\ take a look."\\ Clint makes a \textbf{perception check}. 16\end{tabular}                                                                                                  \\ \hline
Vi                 & \begin{tabular}[c]{@{}l@{}}I'll help as well.  I got a 10\end{tabular}                                                                                 \\ \hline
DM                 & \begin{tabular}[c]{@{}l@{}}Clint, you notice a few goblins crouching in a part \\ of the shaded woods off to the side of the road... \end{tabular}                                                                                                                                                                                    
\end{tabular}
} 
\vspace{-0.2cm}
\caption{ \small Example dialogue transcript from D\&D game play.
}
\vspace{-0.3cm}
\label{tab:example_dialogues}
\end{table}

\subsection{\taskWithEmoji Task}\label{sec:2.2}
Consider three variables in communication between a teacher and a student: \emph{context} $\mathcal{C}$, \emph{teacher utterance} $\mathcal{T}$, and the subsequent \emph{student utterance} $\mathcal{S}$. 
In standard dialogue response generation (RG) setup, models are trained to generate the next utterance only based on the \emph{previous} dialogue history, \emph{i.e.}, $P(\mathcal{T}|\mathcal{C})$ for teacher and $P(\mathcal{S}|\mathcal{C}, \mathcal{T})$ for the student. 
In our task setting, we further consider one variable: \emph{intents} of the teacher: $\mathcal{I_T}$.\footnote{Students also have intents, which are not explicitly modeled in this work.}
In \task, we assume that the teacher's intents are to guide the student to perform certain \emph{action} $\mathcal{A}$ and the intents are fulfilled if the student's subsequent utterance $\mathcal{S}$ entails $\mathcal{A}$.
Since we focus on verbal communication, all variables including $\mathcal{I_T}$ and $\mathcal{A}$ are in natural language (NL).
The teacher model's goal is thus to first come up with an intent, \emph{i.e.}, $P(\mathcal{I_T}|\mathcal{C})$ and then generate an utterance that helps achieve the intent, \emph{i.e.}, $P(\mathcal{T}|\mathcal{C}, \mathcal{I_T})$ such that $\mathcal{S} \approx \mathcal{A}$, given student model $P(\mathcal{S}|\mathcal{C}, \mathcal{T})$. 


 \subsection{D\&D Dialogue Generation as a Partially Observable Markov Decision Process} \label{sec:2.3}
Here we discuss a reformulation of the standard RG problem as a partially observable Markov decision process (POMDP).
We consider a POMDP defined as $\left<S, A, T, R, O\right>$, where $S$ is a set of states, $A$ is a set of actions performed by the teacher (note it is different from the player action $\mathcal{A}$), $T$ is a set of transition probabilities between states ($T(s'|s, a)$), $R$ is reward function, and $O$ is a set of observations.
In D\&D dialogues such as Table~\ref{tab:example_dialogues}, we consider the first DM sentence (not in bold) as the \emph{observation} containing an incomplete description of world \emph{state}, the second sentence in bold as the \emph{action} containing guidance for players, the next player turns as \emph{reward} (in this case players' perception check\footnote{ \emph{Ability check} is a game mechanic that models the stochasticity in D\&D. The player needs to roll a die and the number determines whether the action succeeds or not.} matches DM intent), and the final turn as new \emph{observation}.
\begin{figure}[tb]
	\centering
	\includegraphics[width=\columnwidth]{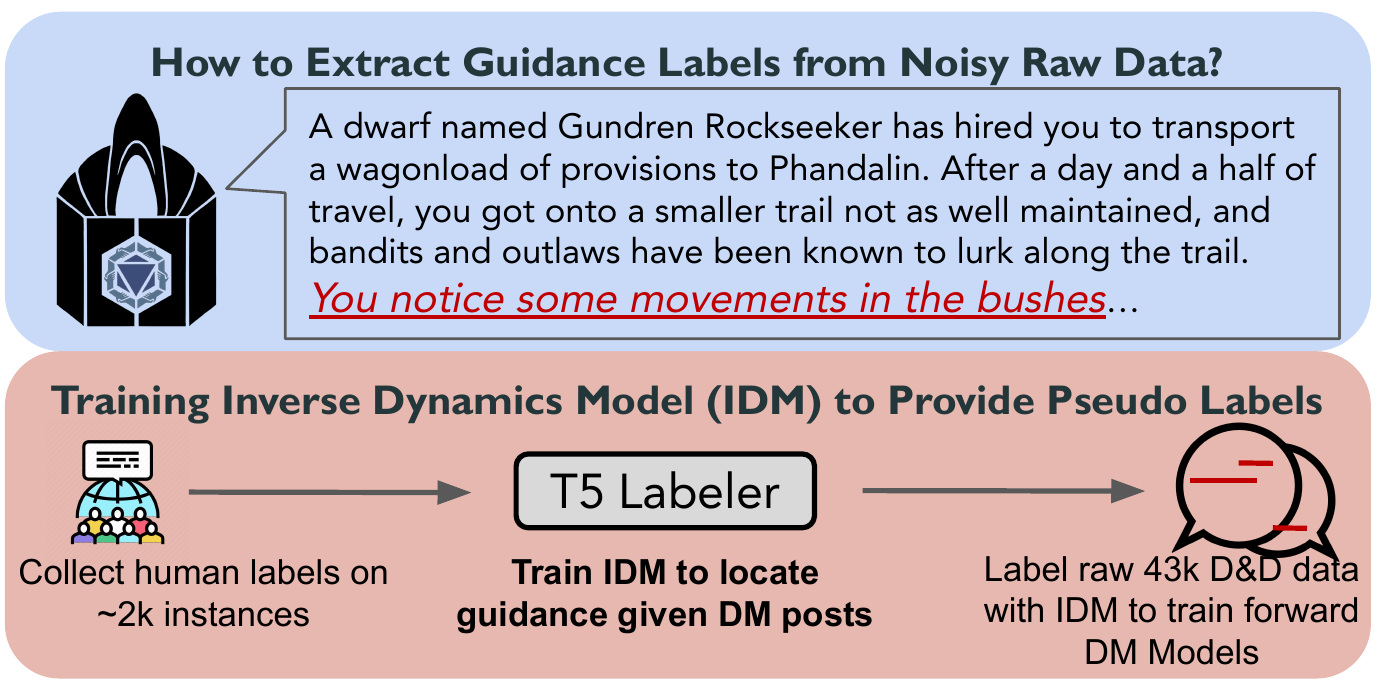}
	\caption{
 \small
	{Illustration of IDM. We collect 2.5k human labels on guidance and train an IDM labeler to generate pseudo labels for unlabeled large corpus. }}
	  	 \vspace{-0.5cm}
	\label{fig:IDM}
\end{figure}
\subsection{Play-By-Post D\&D Data}\label{sec:2.1}


As introduced in Sec.~\ref{intro}, D\&D satisfies two crucial aspects we investigate in \task: \emph{goal-driven} (players are motivated to finish quests guided by the DM) and \emph{groundedness} (players and DM are co-located in the environment and narratives).
Furthermore, the DM is constantly providing guidance to other players, matching the \emph{teacher} role in \task.
We use \emph{actual play} game transcript dataset from~\citet{callison-burch-tomar-et-al-2022} scraped from Play-By-Post (PBP), a web forum\footnote{\tiny \url{https://www.dndbeyond.com/forums/d-d-beyond-general/play-by-post}} where people play D\&D by taking turns posting on the forum. 
PBP data contains more than 800k turns with around 58M words, annotated heuristically with game state information such as player class, race, and ability checks. 
However, to adapt this dataset to our \task setting, we need to filter the data to focus on interactions of DM guiding players. 
Details are in Appendix~\ref{appendix: cleaning detail}.

\subsection{Creating the Environment}\label{sec:2.4}
Training a DM to generate guidance using \task formulation requires first identifying which part of DM's utterances contains guidance, as the DM also roleplays other characters, chitchat, or discusses rules.
Creating such labels requires human-in-the-loop data collection or large offline labeled datasets, both of which are heavily resource intensive~\cite{fu2020d4rl}.
To mitigate such resource constraints, we collect human labels on a small ($<5\%$) portion of our dataset and then train an inverse dynamics model (IDM) that given the players' reactions (\emph{reward} $R$) after potential DM guidance (\emph{action} $A$), extracts which portions of the DM's utterance contain guidance (Figure~\ref{fig:IDM}).

Given that we cast the dialogue generation in \task as a POMDP, the \emph{forward} modeling problem is to generate guidance so that the player's feedback is as intended, such as \emph{making a perception check}. 
Thus our \emph{inverse} modeling problem can be formulated as given the next player ability check being \emph{perception check} (feedback/reward), extracting the guiding sentence (DM's action) from DMs' utterances. 
IDM modeling is simpler than forward behavior cloning because it uses a non-causal formulation that exploits both past and future events to identify a guidance sentence~\cite{baker2022video}. 

\textbf{Human Label Collection.} 
We design our human labeling interface to contain 3 questions: 1. \emph{Does this DM turn contain guidance or not?} 2. \emph{If it does, please choose a sentence from the text that serves the purpose of guidance the most.} 3. \emph{Imagine that you were the player, what ability check would you make?} We add the third question to provide more labels to evaluate DM models (discussed in Section~\ref{sec:4.3}). Details are in Appendix~\ref{appendix:human label details}.

\textbf{IDM Training.}
In practice, we collect around 2.5k human labels on guidance and train IDM to provide labels for the large unlabeled data. 
We consider two subtasks for IDM: \emph{identifying} whether a DM turn (DT) contains guidance and \emph{extracting} the key guiding sentence (GS) from DT.
We train two T5-3B models~\cite{raffel2020exploring}, one for classifying DM texts that contain guidance or not (\emph{IDM-Identify}) and the other for extracting a sentence from the text (\emph{IDM-Extract}). More details can be found in Appendix~\ref{appendix:IDM detail}.

\textbf{IDM Results.}
We evaluate IDM performance on 1k human-labeled data and compare it to baselines such as the longest sentence and GPT-3 with in-context learning. Detailed results are in Appendix~\ref{appendix:IDM detail}. In summary, we find that trained IDM outperforms other baselines on extracting GS, reaching around 70\% accuracy where random guessing is 10\% (the average number of sentences in DM's posts is around 10).


\section{Theory-of-Mind Inspired Guidance Generation in Grounded Environments} \label{modeling}

\begin{figure}[tb]
	\centering
	\includegraphics[width=0.9\columnwidth]{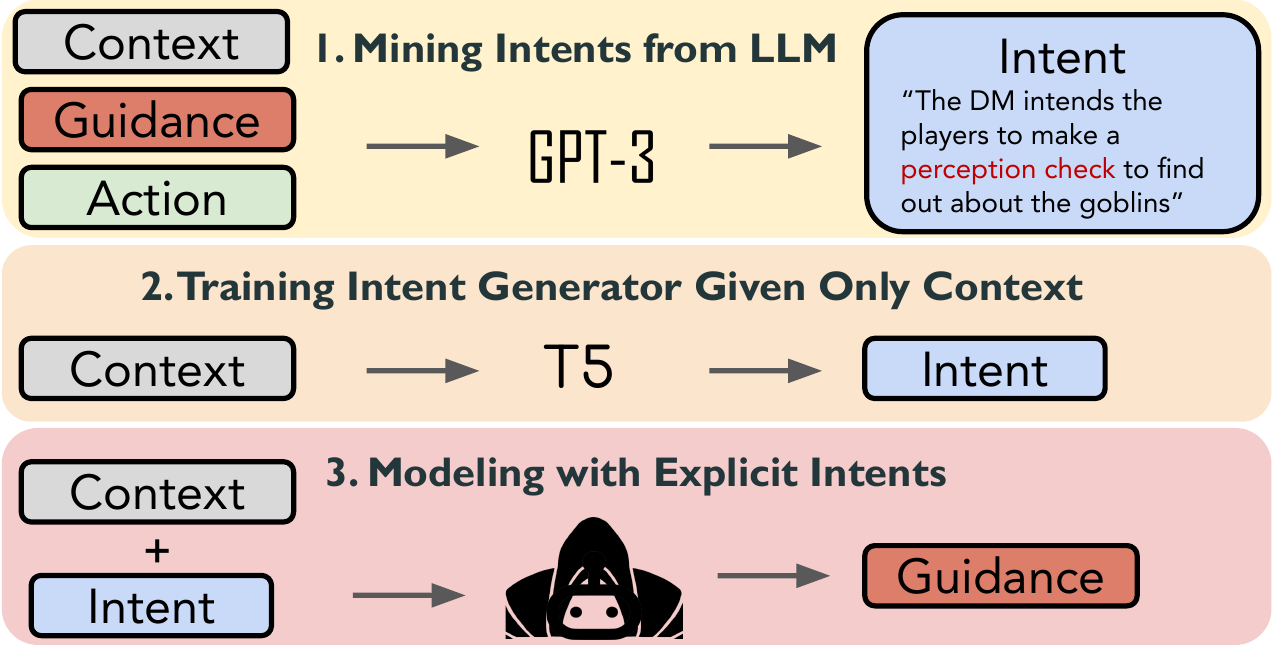}
	\caption{
 \small
	{Illustration of intent modeling. We first mine intents from LLM and then train an intent generator to generate intent as additional context to train the DM model.}}
	  	 \vspace{-0.4cm}
	\label{fig:intent_procedure}
\end{figure}

This section introduces our exploration of model designs to train a teacher model that can guide the student to perform certain actions by speaking in a grounded environment.
We are specifically interested in the research question ``\emph{Does incorporating \textbf{intent} (\ref{sec:3.1}) and \textbf{theory-of-mind} (\ref{sec:3.2}) help models generate better guidance?}'' 

\subsection{Modeling Intents}
\textbf{\emph{Implicit} Intent.}\label{sec:3.1}
We start with the standard RG setup in most dialogue modeling work: training models to directly generate the target utterance (guidance) given dialogue context with no explicit intent involved. Formally, we model $P(\mathcal{T}|\mathcal{C})$ using the DM text with guidance as teacher target utterance $\mathcal{T}$ and the context turns as $\mathcal{C}$.

\textbf{\emph{Explicit} Intent with Generator.}
Here we propose modeling methods that include explicit intents of the teacher $\mathcal{I}_T$. 
Following~\ref{sec:2.2}, we treat the teacher's intents as additional context appended to the dialogue context, \emph{i.e.}, $P(\mathcal{T}|\mathcal{C}, \mathcal{I_T})$. Figure~\ref{fig:intent_procedure} shows the procedure.
\textbf{1. Mining Intents Using Large Language Models (LLMs)}
Since intents are implicit in the data, we first need to mine DM's intents from their utterances. 
To ensure the quality of mined intents, we use LLM such as GPT-3 to generate intents in natural language given context, guidance sentence from DM, and the next-turn player action. 
We prompt GPT-3\footnote{We use text-davinci-03 from~\url{https://beta.openai.com/docs/models/gpt-3}} with ``\emph{The following is a conversation that happened in a game of Dungeons and Dragons: [Context] [DM Text] [Player Name]:[Player Ability Check] Question: What do you think that the DM intentds to do by mentioning [Extracted Guiding Sentence]? Answer:}'' 
\textbf{2. Training Intent Generator} 
Using mined intents, we train an \emph{intent generator} (IG) that takes the context $\mathcal{C}$ as input and generates an output of the DM's potential intent $\mathcal{I}_T$. 
In practice, we train a sequence-to-sequence model T5~\cite{raffel2020exploring} on 45k mined intents for our training and valid data. 
We also conduct a human evaluation on both mined and generated intents to examine whether these intents are reasonable given the context. Humans rate 85\% of the mined intents and 75\% of generated intents proper with 3-way redundancy of each intent from sampled 500 intents.
\textbf{3. Modeling with Generated Intent} With a trained IG, we then generate intents on our test split. Then the teacher model that takes intents as additional input will use the generated intents from IG to generate utterances during testing. 


\begin{figure*}[t!] \begin{center}
    \includegraphics[width=0.9\linewidth]{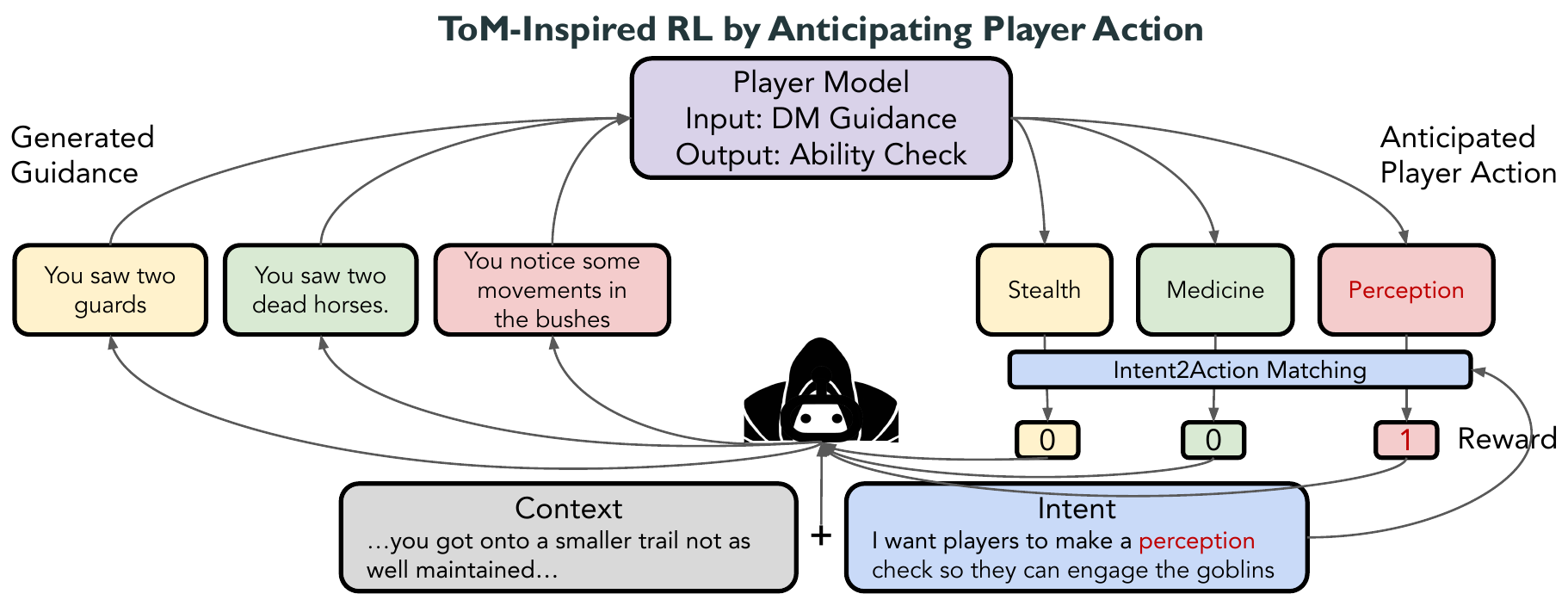}
    \caption{ \small
        Illustration of our ToM-Inspired RL by using a reward function to help DM model anticipate what the players might do upon hearing the generated guidance. We give the model a reward if the predicted player action matches the intent given.}
    \label{fig:ToM_diagram}
\end{center} 
\vspace{-0.6cm}
\end{figure*}

\subsection{Modeling (Limited) \emph{Theory-of-Mind} (ToM) Using RL for Guidance Generation}\label{sec:3.2}
\textbf{Background and Intuition.}
Here we model a limited scope of ToM by modeling the anticipated action of the players in order to help the teacher to generate utterances that guide students to fulfill the teacher's intents. 
Specifically, in \dataWithEmoji, the DM infers what the players might do when they provide different guidance. 
For example, ``\emph{you notice some movements in the bushes}'' will likely motivate the players to make a \emph{perception} check while ``\emph{the guard seems a bit shaken to hear your words}'' might prompt the players to make a \emph{persuasion} check. 
DM then chooses the guidance that will more likely prompts players to perform the action that fulfills the goal.

\textbf{Training Player Model.}
The first step of our proposed ToM-teacher is to train a \emph{player model} (PM) that takes in context and DM utterances and outputs the most likely player action (ability check), \emph{i.e.}, $P(\mathcal{A}|\mathcal{C}, \mathcal{T})$. 
Luckily, each instance of our \data data naturally contains training data for PM with the DM turn and next-turn player ability check. 
We also train a sequence-to-sequence model T5~\cite{raffel2020exploring} to predict the player action using our data. 
The trained PM reaches around 71\% accuracy in predicting the actual player ability check. 
To get an approximate upper bound of the task, we ask humans to predict the next player action on our test set and observe only about 76\% accuracy in matching with players in the data transcript. 
This might be due to the players actually playing the game also considering other factors when making the decisions that we do not have in our data: long-term character goal, detailed persona, player roleplaying style, etc.
We argue that our player model presents a reasonable proxy of what a player might act given the context provided.

\textbf{Player Action-Intent Matching as Reward.}
With a player model approximating player reactions, we then use Reinforcement Learning (RL) to reward the DM model if it generates guidance that will lead the PM to perform an action matched with intent (Figure~\ref{fig:ToM_diagram}). 
Specifically, during training the \emph{Mined Intent} and \emph{Generated-Intent} models introduced in Section~\ref{sec:3.1} to model $P(\mathcal{T}|\mathcal{C}, \mathcal{I_T})$, we pass the model output $\mathcal{T}$ to the trained PM ($P(\mathcal{A}|\mathcal{C}, \mathcal{T})$) and get predicted player action $\mathcal{A}$. 
Since intents are in NL, we train a matching module \emph{Intent2Action} to convert them to the most likely ability check such as ``\emph{perception}'' (23 types in total), $P(\mathcal{A_T}|\mathcal{I_T})$.
Finally, we examine whether the predicted action from PM ($\mathcal{A}$) matches with the \emph{intended} action (ability check) from the DM ($\mathcal{A_T}$). 
Finally, we give the model reward of 1 if the actions match and 0 if not. 
Intuitively this helps shape models to generate guidance more aligned with intents by simulating what the players might do one step ahead.


\section{Evaluating \taskWithEmoji} \label{eval}



Here we propose multifaceted evaluation protocols to measure the quality of the DM/teacher model for \task.
We introduce three criteria, \textbf{Fluency, Groundedness, and Goal-Fulfillment}, to evaluate model outputs.
We design automatic metrics and human evaluation protocols for each criterion, and analyze how well the proposed metrics correlate with human judgments in~\ref{sec:5.2}.
We refer to outputs satisfying all three criteria as \textbf{\emph{star DM}}.

\subsection{Measuring Fluency}\label{sec:4.1}
We first examine whether the output text sounds natural and fluent as a DM. 

\textbf{Automatic Metrics: Matching with References.} 
As with most dialogue evaluation metrics, we use human-written responses as ground truth references and compare the output with them. 
The closer the output is to the human original response, the more fluent\footnote{Perplexity is also often used to measure fluency, but this measure isn't relevant as we don't use autoregressive LMs.}.
We use standard natural language generation (NLG) metrics such as BLEU~\cite{papineni2002bleu} and ROUGE~\cite{lin2004rouge}, etc. to measure the overlap between the output and reference.

\textbf{Human Evaluation.}
For each response, we ask three annotators to ``\emph{evaluate whether the response sounds natural and fluent. If anything seems off or sounds weird—confusing, illogical, repetitive, or factually wrong—then choose No.}'' and use majority voting.


\subsection{Measuring Groundedness}\label{sec:4.2}
\task focuses on grounded communication, where the teacher and student share environment and background knowledge. 
Thus, here we focus on evaluating whether the generated output is \emph{grounded} to the context of the story built by the DM and players.

\textbf{Automatic Metrics: Entity Matching.}
We design an automatic metric to measure the \emph{entity overlap} between those mentioned in the context and in the generated output. 
Intuitively, the generated responses should not have mentions of entities not in the context, otherwise, the model is hallucinating. 
We use a RoBERTa-large-based~\cite{liu2019roberta} named entity recognizer (NER) to extract entity mentions such as person's names and locations from both the context and the model output and calculate their overlap (the higher the better).

\textbf{Human Evaluation.}
Since groundedness also covers other aspects (narrative flow, style, etc.) than entities, we conduct a human evaluation to measure whether the response sounds like it is continuing the same story from context. 
For each response, we ask three annotators to ``\emph{evaluate that given the conversation context, whether the response sounds like it's continuing the same story (grounded), or beginning a new story (NOT grounded)?}''

\begin{table}[tb]
\centering
\resizebox{\columnwidth}{!}{
\begin{tabular}{clll}
\hline
\multicolumn{2}{c}{\textbf{Model Variant}}                                                                                             & \textbf{Base Model} & \textbf{Input} \\ \hline
\multicolumn{1}{c|}{\multirow{2}{*}{Implicit Intent}}                                                                & Human-Label     & T5-3B               & Context        \\
\multicolumn{1}{c|}{}                                                                                                & IDM-Label       & T5-3B               & Context        \\ \hline
\multicolumn{1}{c|}{\multirow{2}{*}{Explicit Intent}}                                                                & Mined Intent    & T5-3B               & Context+Intent \\
\multicolumn{1}{c|}{}                                                                                                & Gen. Intent     & T5-3B               & Context+Intent \\ \hline
\multicolumn{1}{c|}{\multirow{2}{*}{\begin{tabular}[c]{@{}c@{}}Explicit Intent + \\ ToM-Inspired RL\end{tabular}}} & RL+Mined Intent & T5-Large            & Context+Intent \\
\multicolumn{1}{c|}{}                                                                                                & RL+Gen. Intent  & T5-Large            & Context+Intent \\ \hline
\end{tabular}
} 
\caption{ \small Model variants. All targeted outputs are guidance from DM. All training data size is 41k except for human-label (2k). The test set (1k) is shared across all.
}
\vspace{-0.5cm}
\label{tab:models}
\end{table}
\begin{table*}[tb]
\centering
\small
\resizebox{\linewidth}{!}{
\begin{tabular}{cc|c|cc}

\multicolumn{1}{c|}{\textbf{Dimensions}} & \textbf{Metrics}                                                            & \textbf{Human-Label 2.5k} & \textbf{IDM-Label 41k}              & \textbf{Random-Label 41k} \\ \hline
\multicolumn{1}{c|}{\textbf{Fluency}}                           & Human Evaluation                   & 0.80                      & \multicolumn{1}{c|}{\textbf{0.81}}  & 0.56                      \\ \hline
\multicolumn{1}{c|}{\multirow{2}{*}{\textbf{Groundedness}}}     & Entity Matching                    & 0.749                     & \multicolumn{1}{c|}{\textbf{0.776}} & 0.718                     \\
\multicolumn{1}{c|}{}                                           & Human Evaluation                   & 0.91                      & \multicolumn{1}{c|}{\textbf{0.92}}  & 0.72                      \\ \hline
\multicolumn{1}{c|}{\multirow{5}{*}{\textbf{Goal-Fulfillment}}} & Guidance Classification            & 0.438                     & \multicolumn{1}{c|}{\textbf{0.474}} & 0.254                     \\
\multicolumn{1}{c|}{}                                           & Player Action Matching   & 0.261                     & \multicolumn{1}{c|}{\textbf{0.262}} & 0.249                     \\
\multicolumn{1}{c|}{}                                           & Human Evaluation - Guidance        & 0.21                      & \multicolumn{1}{c|}{\textbf{0.23}}  & 0.20                      \\
\multicolumn{1}{c|}{}                                           & Human Evaluation - Action Matching & 0.11                      & \multicolumn{1}{c|}{\textbf{0.17}}  & 0.13                      \\ 
\end{tabular}
} 
\caption{\small Results on the 3 dimensions using metrics from Section~\ref{eval} comparing models that use IDM-generated pseudo-labels and human-generated labels.
}
 \vspace{-0.4cm}
\label{tab:IDMAuto}
\end{table*}
\subsection{Measuring Fulfillment of Intents}\label{sec:4.3}
The core measure of the success of models for \task is whether the goal of the teacher is fulfilled by making the response. Specifically, we want to measure, whether the generated output 1) indeed contains \textbf{\emph{guidance}} for the student and 2) guides the student to perform the action that the teacher wants them to take (\textbf{\emph{action matching}}).

\textbf{Automatic Metrics: Guidance Classifier and Player Action Matching.}
To evaluate whether the generated output contains any \textbf{\emph{guidance}}, we reuse the \emph{IDM-Identify} model discussed in~\ref{sec:2.1} that takes the input of DM posts and predicts whether this post contains guidance or not.
For \textbf{\emph{action matching}},
since it is infeasible to collect the original players' responses on all model outputs, we train a player model (PM) to generate potential actions given DM model outputs. 
Finally, we compare the predicted action with the actual player action after the human DM guidance from the dialogue transcript.
The higher the percentage of matching human player action, the better the model is at generating guidance that achieves the same goal as human DM.
Note that although we also train a PM for ToM modeling in~\ref{sec:3.2}, the PM used for evaluation is a distinct model based on a larger model and trained on the test set of the data as well.

\textbf{Human Evaluation.}
To evaluate \textbf{\emph{guidance}}, we ask annotators: ``\emph{Is this response providing guidance to the players?}'' 
For \textbf{\emph{action matching}}, we ask crowdsourcing workers to write down the most likely ability check that they think the player will take after the given DM utterance. 
We also provide annotators with the player character's race and class to better approximate the players.




\section{Experimental Results} \label{exp_results}

We aim to answer three research questions through our experiments: 1) \emph{Do IDM-provided labels help train models that generate better guidance?} 2) \emph{Does explicitly incorporating intents result in better models?} 3) \emph{Does theory-of-mind modeling help models become better communicators?}
\subsection{Compared Models}
We use T5-3B~\cite{raffel2020exploring} as our base model. We train a model with only 2.5k human-labeled guidance data collected in~\ref{sec:2.4} (\textbf{Human-Label}). 
Then we train IDM on human labels and provide labels for the rest of the 41k unlabeled dialogues (\textbf{IDM-Label}). 
Next, we explicitly incorporate intents in modeling and consider two model variants following~\ref{sec:3.1}: \textbf{Mined Intent} that is given intents mined from LLM using both context and next-turn player actions; \textbf{Generated Intent}, where the model is trained on mined intents, but during test time, we train an intent generator to provide intents without knowing future turns.
Finally, following Section~\ref{sec:3.2}, we use a trained player model to provide reward signals for DM models for RL. 
We use T5-Large for RL training on top of mined intent (\textbf{RL-ToM-Mined}) and generated intent (\textbf{RL-ToM-Gen.}) models. 
We use RL4LMs~\cite{ramamurthy2022reinforcement} to implement the reward function and use Proximal Policy Optimization (PPO)~\cite{schulman2017proximal} for RL training. 
A summary of model variants is shown in Table~\ref{tab:models}.

\subsection{Correlation Analysis of Automatic Metrics}\label{sec:5.2}
Here we present correlation results of automatic metrics in Sec.~\ref{eval} using human evaluation results (with an average inter-annotator agreement of 0.78) on our test set.
For \textbf{fluency}, we find a statistically insignificant correlation (p-values $>0.05$) between automatic metrics that measure lexical matching with a reference response.
We suspect that 1) lexical matching does not reliably capture the naturalness of languages~\cite{sagarkar2018quality, delucia2021decoding} and 2) many plausible responses can be made given the same context~\cite{zhou2022reflect}, making comparing with the single reference unreliable. 
For both \textbf{groundedness} and \emph{goal-fulfillment}, we find statistically significant (p-value $<0.0001$) correlations between automatic metrics (entity matching, guidance classifier, and action matching) and human judgments on test instances. 
\textbf{Conclusion}: for \textbf{fluency}, we will use human evaluation and for \textbf{groundedness} and \textbf{goal-fulfillment}, the automatic metrics provide a reasonable proxy.

\begin{figure}[tb]
	\centering
     \vspace{-0.3cm}
	\includegraphics[width=1.0\columnwidth]{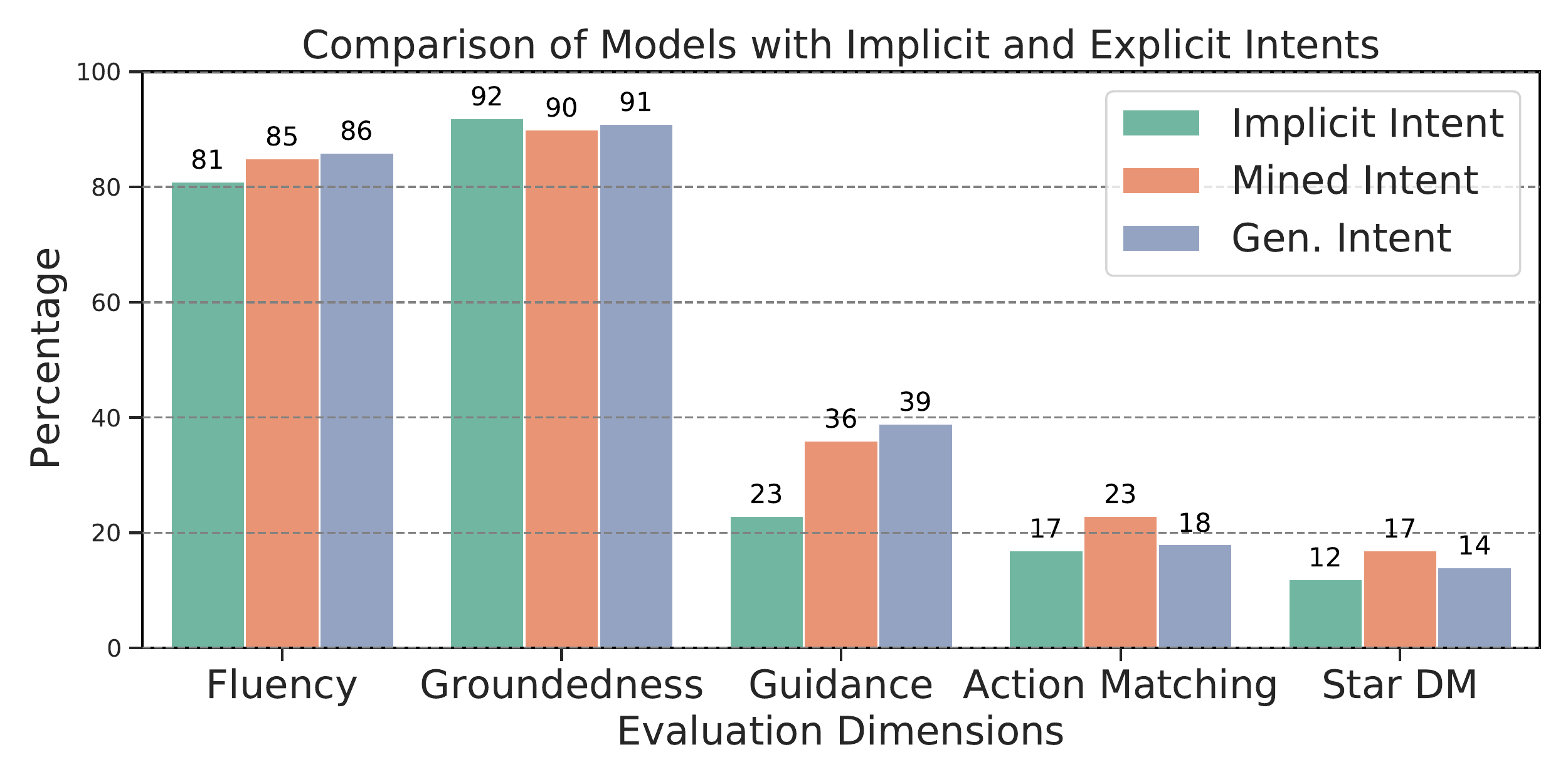}
	\caption{ \small Results comparing implicit and explicit intent models. We observe models with intent generate dramatically more guidance.
	}
\vspace{-0.5cm}
	\label{fig:intent_comparison}
\end{figure}

\begin{figure*}[ht]
\centering
\begin{subfigure}[b]{0.49\textwidth}
\includegraphics[width=\textwidth]{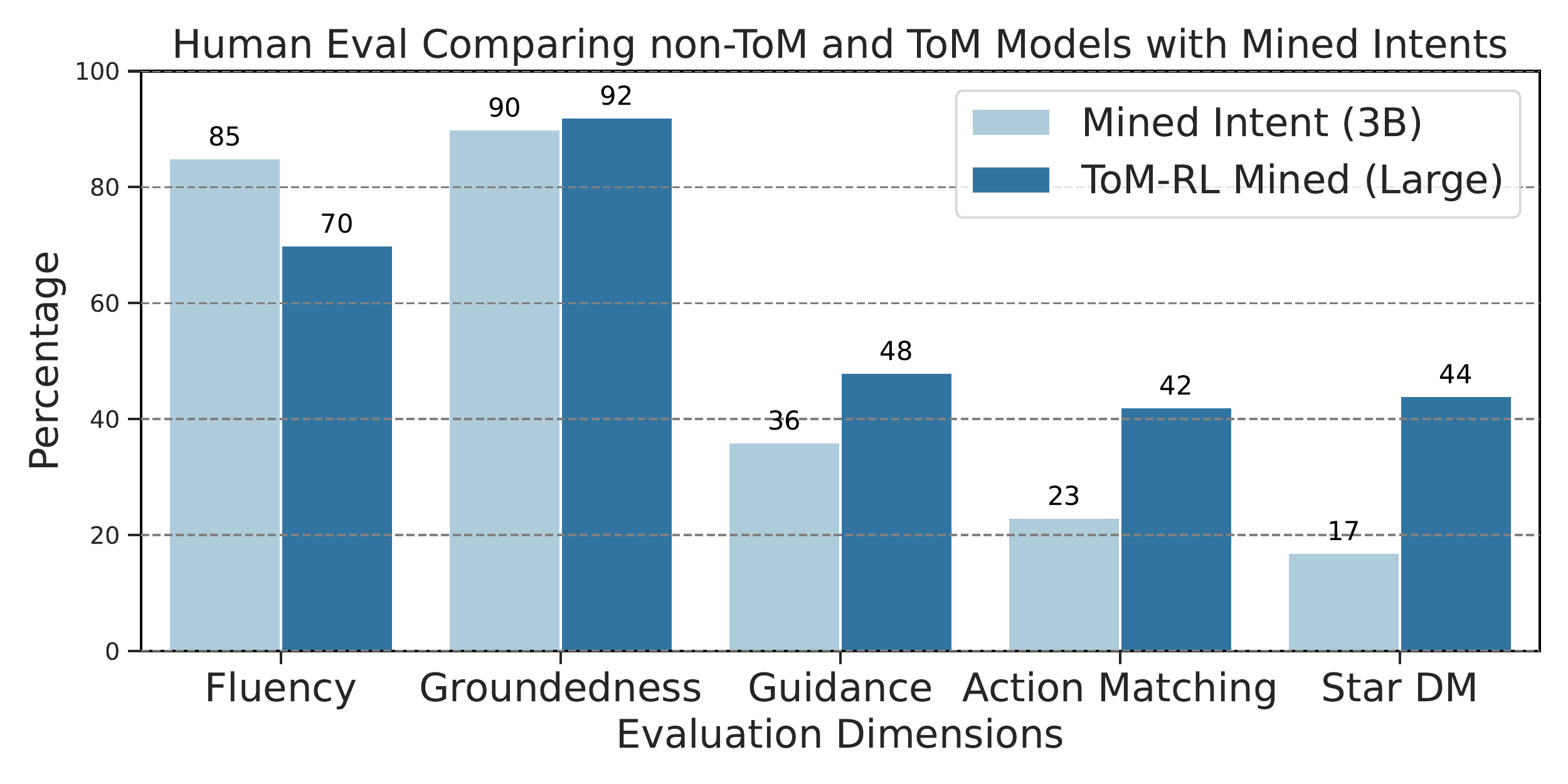}
\end{subfigure}
\begin{subfigure}[b]{0.49\textwidth}
\includegraphics[width=\textwidth]{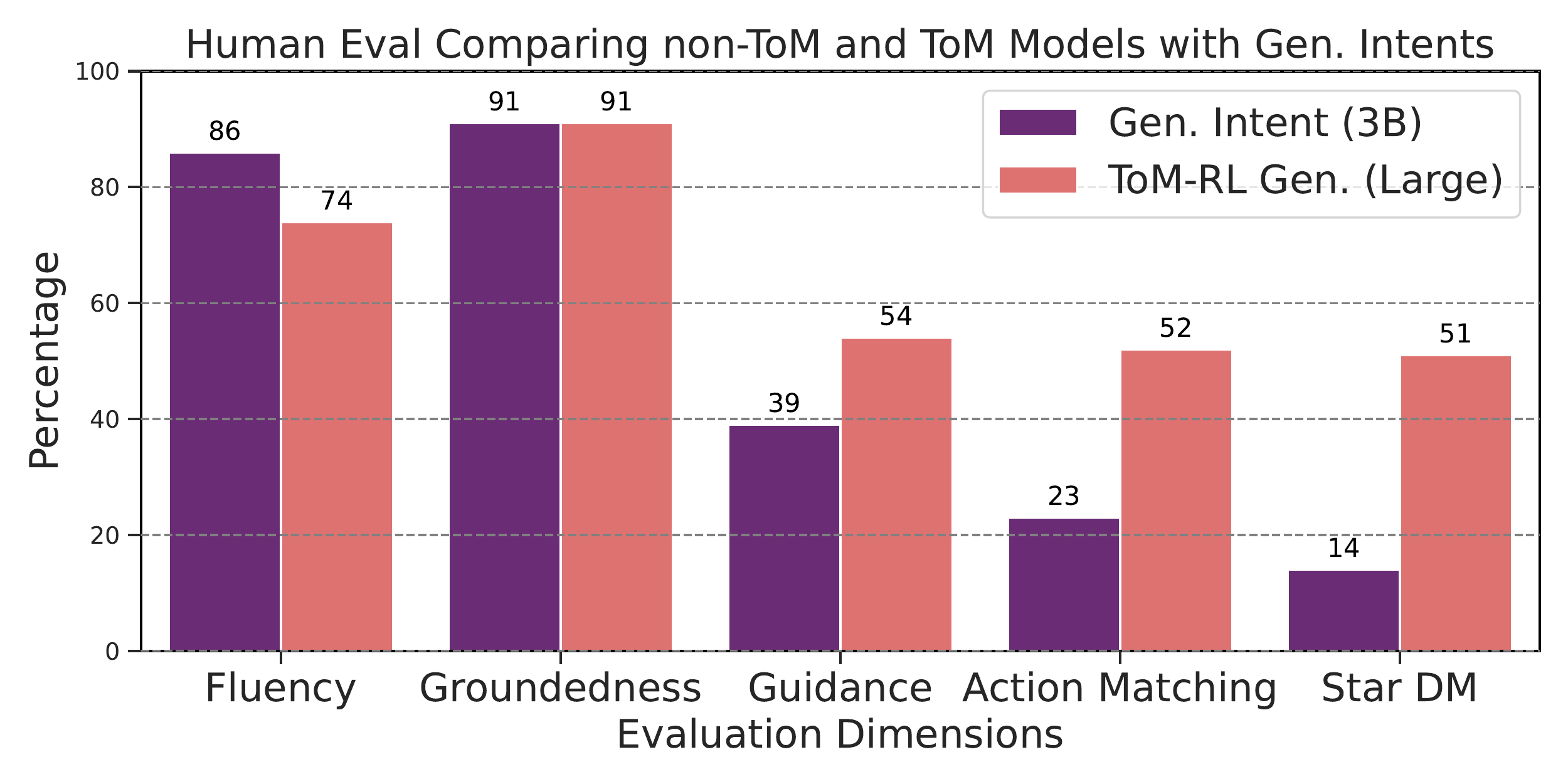}
\end{subfigure}
\caption{ \small
Human Evaluation comparing non-ToM and ToM models with mined (Left) and generated (Right) intents.
}
\vspace{-0.3cm}
\label{fig:ToM_comparison}
\end{figure*}

\begin{figure}[tb]
	\centering
     \vspace{-0.3cm}
	\includegraphics[width=1.0\columnwidth]{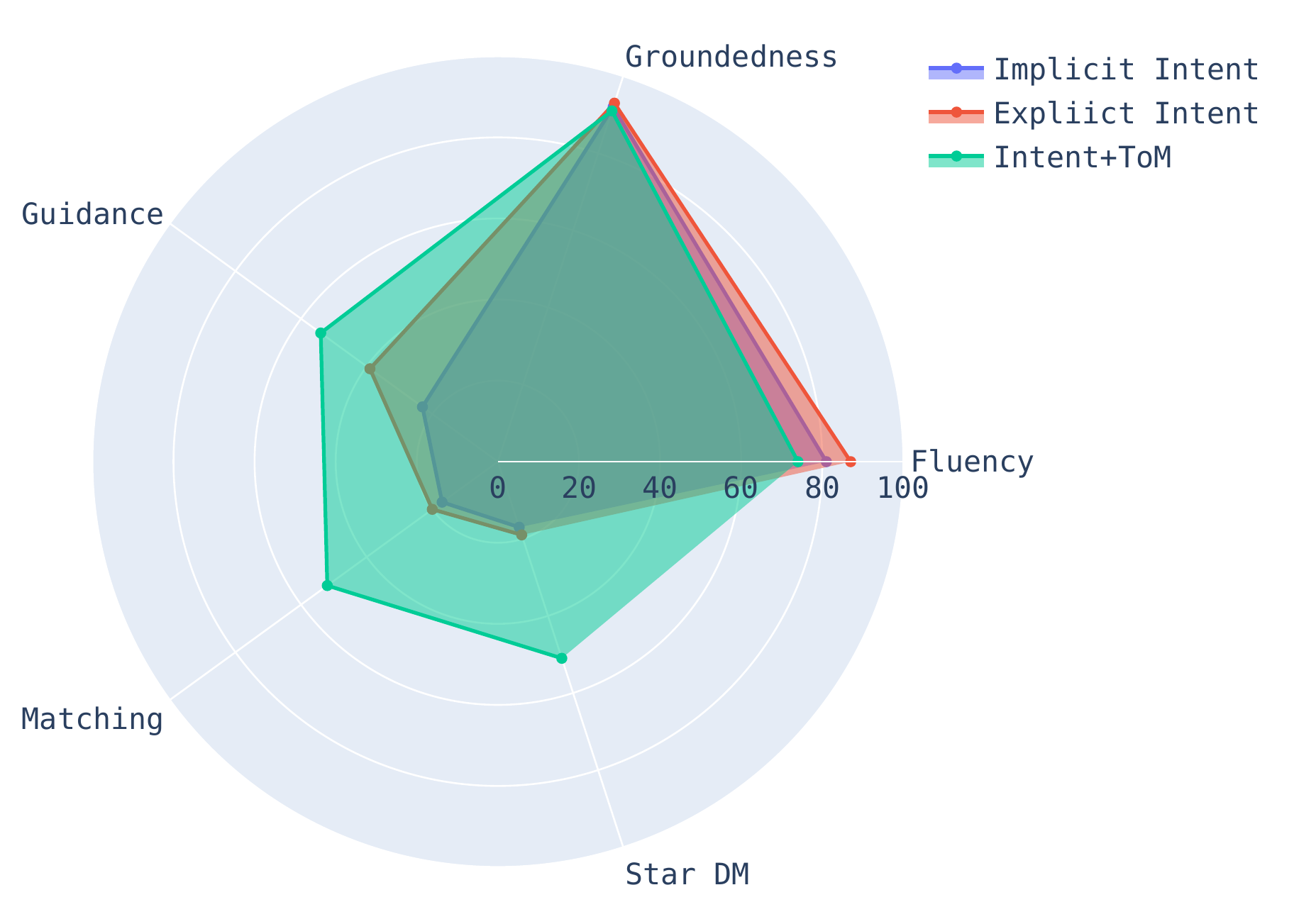}

	\caption{ \small Summary of performance on different evaluation aspects from the BEST 1) implicit intent model (IDM-Label 41k, 2) explicit intent model (Mined Intent), and 3) intent with ToM-inspired RL (ToM-RL Generated Intent).
	}
\vspace{-0.6cm}
	\label{fig:radar}
\end{figure}
\subsection{Results and Analysis}
\textbf{Do IDM-provided labels help models generate better guidance?}
Here we examine the effects of our inverse dynamics models on training DM models for \task. 
Table~\ref{tab:IDMAuto} presents the results following our evaluation dimensions introduced in Section~\ref{eval}. 
We see that models trained using our IDM-provided labels outperform those trained on the small number of high-quality human labels on \emph{all} measures. 
To show that data size alone is not sufficient for training a good DM model, we randomly assign labels of guiding sentences on the same number of training instances as IDM models (``\emph{Random-Label 41k}'') and find the performance is significantly worse than either of the models using human or IDM labels. 
This shows that the quality of IDM-provided labels is critical for DM modeling and our IDM offers a scalable and affordable solution to obtain a large number of quality labels requiring only small-scale human annotation.

\textbf{Does explicitly incorporating intents help?}
Figure~\ref{fig:intent_comparison} shows results comparing the best model with no explicit intents (IDM-Label), mined intents, and generated intents. 
We find that models with explicit intents perform on par on groundedness, but improve on fluency, guidance, and action matching. 
The improvement is especially dramatic on the \emph{Goal-Fulfillment} aspect, as adding intents increases the proportion of outputs that contain guidance by more than 50\% and action matching by more than 30\%. 
We speculate that this might be due to explicit intent modeling, as the model is biased towards generating output that is aligned with the intent instead of purely modeling the most likely next possible sequence of tokens.


\textbf{Can we model theory-of-mind using reinforcement learning?}
Last but not least, we are interested in whether the ToM-inspired reward function we design in Section~\ref{sec:3.2} can help train better communication models for \task. 
Figure~\ref{fig:ToM_comparison} shows the results of adding ToM to mined intent (left) and generated intent (right) models. 
We find that despite using a much smaller (1/4 parameter) base model, models with reward function mimicking ToM can outperform the no-ToM variants on generating 40\% more outputs with guidance that lead to players to perform the action matching intents while performing on par on groundedness. 
We also find that the fluency drops, possibly due to using a smaller base LM (due to memory constraints) and RL training affects the naturalness of outputs. 
Potential remedies we plan to explore in the future include using larger models and modifying the reward function to also account for fluency such as using KL divergence.
Even with the drop in fluency, however, we still observe that with ToM-inspired RL, models can generate responses that satisfy \emph{all} measures (star DM) up to 3.5 times more than without ToM modeling. 

Finally, we present an overall comparison between the \emph{best} models under each category (implicit intent, explicit intent, explicit intent with ToM modeling) in Figure~\ref{fig:radar}. All three variants perform on par with groundedness. And while fluency drops when adding explicit intents and ToM, these two additions improve dramatically on the goal-driven aspects (guidance and action matching). Models with both explicit intents and ToM modeling using RL perform overall the best and produce almost threefolds of human DM-like (star) responses than others. This shows a promising sign that both intents and ToM-inspired RL can help goal-driven models to better achieve their communicative intents.


\section{Related Work}\label{rel_work}
\textbf{Goal-Driven Grounded Dialogue Agents.}
There is an emerging line of works studying goal-driven situated dialogues~\cite{urbanek2019light,narayan2019collaborative, ammanabrolu2021motivate, bara2021mindcraft,prabhumoye2020love,padmakumar2022teach,ammanabrolu2022dialogue}. 
However, intents or ToM are rarely incorporated explicitly in developing more human-like communication agents. 
CICERO~\cite{meta2022human} proposes a strategy-guided dialogue generation agent to play Diplomacy with modeling other players' next moves. 
We argue that most prior work along this line (text games, Diplomacy) is still a more constrained set of scenarios compared to D\&D.

\textbf{Dungeons and Dragons as an NLP Challenge.}
Several studies have used Dungeons and Dragons to study various problems in NLP such as character understanding~\cite{louis2018deep}, controlled dialogue generation~\cite{si2021telling, callison-burch-tomar-et-al-2022}, and description generation~\cite{newman2022generating}. Reinforcement learning has also been applied to study the goal-driven aspect of D\&D~\cite{martin2018dungeons}. 

\textbf{World Modeling and Simulation.}
D\&D involves world-building and modeling actions which inspires inverse dynamics modeling. A line of work has studied world modeling, generation, and using IDM to create labels for model learning~\cite{ammanabrolu2021learning,ammanabrolu2022dialogue,baker2022video}. Theater script co-writing has also been studied recently~\cite{mirowski2022co} for the simulation of a small-scale world.

\textbf{Theory-of-Mind and Pragmatics.}
Theory-of-mind has been studied in psychology and cognitive science for decades. 
Rational Speech Act (RSA) framework studies pragmatics between speakers and listeners using a probability perspective~\cite{frank2012predicting, goodman2016pragmatic}.
\citet{shafto2014rational} has shown that teaching by simulating the student increases effectiveness.
Recent work has looked into ToM and pragmatics as an essential aspect of language usage~\cite{nematzadeh2018evaluating, le2019revisiting,pu2020program,fried2022pragmatics, sap2022neural}, especially communication~\cite{zhu2021few, bara2021mindcraft}. 

\section{Conclusion}\label{conclusion}
We propose \taskWithEmoji to study goal-driven and grounded language interactions focusing on generating guidance from the teacher to lead students to perform certain actions. 
We use D\&D as our test bed and construct large-scale data \dataWithEmoji by using IDM to provide quality labels. 
We train models to generate guidance by modeling intents and theory-of-mind. 
Results show a promising sign that incorporating explicit intents and ToM modeling makes better communication agents.

\section{Ethics and Broader Impact}
Our study is conducted in English, which benefits English speakers more. 
D\&D is also more popular in the western world. 
We use Amazon Mechanical Turk to recruit crowdsourcing workers and we pay workers over \$15/hour on average, well above the highest state minimum wage, and engage in constructive discussions if they have concerns about the process. 
We also give each annotation instance enough time so that we do not pressure annotators.

The online forum D\&D gameplay data we use from~\citet{callison-burch-tomar-et-al-2022} might contain aggressive language. 
Our intents are mined from LLM (GPT-3), which might surface or even amplify harmful content within these models, such as biases and private information.
We use a keyword-based filter for both the dialogue and intent data before training our models.

Our work deals with \emph{communicative intents} of neural computational models. However, we want to emphasize that the intents of AI models (especially conversational systems) should be closely monitored and regulated~\cite{crawford2021atlas}.
In our work, we choose a fantasy domain with a relatively low stake to study model intentions with the overall goal of \emph{assisting} players (humans or AI) to have a better experience in a role-playing game.


\section{Limitations}
Here we discuss several limitations of our work and point to potential future work directions. 
First, we focus on single teacher and single student setup to study guidance generation whereas in real life there often are multiple teachers and students. We plan to extend to multi-party goal-driven communication and D\&D also provides a proper testbed to study this problem.

Second, there are more nuances in guidance: railroading direct guidance (``\emph{make a persuasion check}'') and subtle indirect guidance (``\emph{the guards seem to be a bit shaken}''). We did include them in our human labeling and evaluation interface but did not specifically distinguish them during modeling.

Third, due to the constraints on input sizes for most LMs, we have to set a context window to study dialogue generation in D\&D. However, both DM and players have a long-term memory about the comprehensive story progression which might influence how they communicate. As a next step, we plan to use summarization models and adventure books as narrative backgrounds to ground our \taskWithEmoji task with a larger world setting.
We include answers to other \textbf{Frequently Asked Questions (FAQ)} in Appendix~\ref{faq}.
\label{limitations}

\section{Acknowledgements}
This research is based upon work supported in part by the DARPA KAIROS Program (contract FA8750-19-2-1004), the DARPA LwLL Program (contract FA8750-19-2-0201), the DARPA MCS Program (contract through NIWC Pacific N66001-19-2-4031), the IARPA HIATUS Program (contract 2022-22072200005), and the NSF (Award 1928631). We thank anonymous reviewers for providing insightful feedback along with members from USC-NLP group, INK and JAUNTS lab.

Approved for Public Release, Distribution Unlimited. The views and conclusions contained herein are those of the authors and should not be interpreted as necessarily representing the official policies, either expressed or implied, of DARPA, IARPA, NSF, or the U.S. Government.

\bibliographystyle{acl_natbib}
\bibliography{custom}

\clearpage
\appendix
\label{appendix}
\section{Frequently Asked Questions (FAQ)}\label{faq}
\subsection{Why only training a DM model to generate guidance instead of everything a DM says?}
A DM needs to do multiple complex language tasks (see~\citet{callison-burch-tomar-et-al-2022} for more analysis) such as world modeling, storytelling, role playing with a persona, judging rules, etc.
And we argue that these span multiple papers or even thesis.
Instead of conflating all kinds of language tasks DM is performing, we focus on the goal-driven aspect of DM: generating guidacne for players to proceed the story.
This task is both critical since human language usage always comes with a purpose~\cite{allwood1976linguistic} and challenging as even LLMs such as ChatGPT~\cite{chatgpt} often lack the ability to produce an utterance that fulfills a communicative intent.
We also argue that with the key capability of generating guidance fulfilling intents, the model can be combined with models with different focus such as storytelling, describing world state, etc. to mimic a human DM.

\subsection{How generalizable is a DM model on other domains?}
D\&D is a specific domain we choose to study \task due to its grounded and goal-driven nature.
We admit it is non-trivial to directly apply a DM model on other domains.
However, we believe that the insights from our modeling approaches attempting to incorporate intents and ToM can generalize to other domains.
Specifically, explicitly including intents in context and using RL to model ToM by anticipating others' reactions can be easily applied in other scenarios.
For example, we can generate intents for a open-domain chatbot such as expressing empathy toward users or make suggestions on an issue the user is facing and using ToM modeling to better generate utterances that achieve those purposes.

\subsection{Where are the data and code?}
All data and code used to train our models including IDM, player models, Intent2Action, intent generator, and DM models are included in the supplementary materials. For more detailed instructions please check README.md in the uploaded materials. We will release the model checkpoints as well upon publication. We hope our open-source efforts help the community develop more exciting communication systems.


\section{Play-By-Post Data Cleaning Details}\label{appendix: cleaning detail}
To use PBP data for \taskWithEmoji, several non-trivial challenges exist. First, posts from DM often contain many non-guidance noises such as out-of-character chitchat, rule discussion, and combat ruling. Second, DM often addresses multiple players and we focus on teacher-student 2-participant interaction in this work (we leave multi-party goal-driven dialogue to future work). Lastly, dialogues from forums are not strictly chronological, meaning that the n-th post might not be responding to the (n-1)-th post due to asynchrony.
Due to the above challenges, we propose our methods to \textbf{filter raw post data to get thread-like dialogues} between the DM and a player that follows chronological order.

We filter PBP data so that each instance contains three components: 1. context/dialogue history (C); 2. DM turn with potential guidance to a player A (DT); 3. player A action turn (PA). To get such thread-like dialogues, we first need to locate which posts contain clear player actions (as feedback to DM's guidance). Luckily, in D\&D, player actions are often clearly indicated by a game mechanic called ``\emph{ability check}'' where the player has to roll a die to determine whether their actions such as perception or stealth succeed or not. This provides clear signals of when the players have taken action.

We thus regard posts that contain players making ability checks as player action turns PA. Then we look at the previous 20 turns to find potential posts with DM guidance (DT) and context (C). We use two annotated tags from PBP the data: ``\emph{name\_mention}'' and ``\emph{reply\_to}'' to locate the DM posts that address the player who makes the ability check. If no posts have been added in the previous 20 turns, we then add the closest turn from the DM that's not replying to another player. After getting DT, we add turns from the player or DM before the DM turn to our context C, completing a three-component thread-like dialogue instance.

\begin{table*}[tb]
\centering
\resizebox{\linewidth}{!}{
\begin{tabular}{c|l|l}
\textbf{Character} & \multicolumn{1}{c|}{\textbf{Game Dialogue}}                                                                                                                                                                                                                                                                                                                                                                                                                                                                                  & \multicolumn{1}{c}{\textbf{Explanation}}                                                                                                                                                                                                                                                                                                           \\ \hline
DM                 & \begin{tabular}[c]{@{}l@{}}A dwarf named Gundren Rockseeker has hired\\ you to transport a wagonload of provisions to\\ the rough-and-tumble settlement of Phandalin...
\end{tabular} & \begin{tabular}[c]{@{}l@{}}
The DM here is providing background for the \\ players and \textbf{sets up an encounter with the goblins}\\, who will provide players with \textbf{important clues.}\end{tabular} \\ \hline \cline{1-2}
DM                 & \begin{tabular}[c]{@{}l@{}}\textbf{You all notice some movements in the bushes nearby} \\ \textbf{the road...}\end{tabular}                                                                                                                                                                                                                                                                                                                                                                                                                    & \begin{tabular}[c]{@{}l@{}}The DM provides \textbf{guidance to prompt players to}\\ \textbf{check surroundings} so that they can find out\\ about the goblins\end{tabular}                                                                                                                                                                  \\ \hline
Clint              & \begin{tabular}[c]{@{}l@{}}"There might be something hiding there, let’s go \\ take a look."\\ Clint makes a \textbf{perception check}. 16\end{tabular}                                                                                                                                                                                                                                                                                                                                                                               & \begin{tabular}[c]{@{}l@{}}The player is making a \textbf{perception check}: a game\\ mechanic that models the stochasticity in the D\&D\\ world. The player needs to roll a die and the number \\ determines whether the ability check succeeds or not.\end{tabular}                                                                    \\ \hline
Vi                 & \begin{tabular}[c]{@{}l@{}}I'll help as well.  I got a 10\end{tabular}                                                                                                                                                                                                                                                                                                                                                                                                                                                     &                                                                                                                                                                                                                                                                                                                                \\ \hline
DM                 & \begin{tabular}[c]{@{}l@{}}Clint, you notice a few goblins crouching in a part \\ of the shaded woods off to the side of the road. \\ Two of the goblins begin charging your wagon...\\ Roll for initiative!\end{tabular}                                                                                                                                                                                                                                                                                                       & \begin{tabular}[c]{@{}l@{}}The Dungeon Master describes the outcome of the\\ perception check and starts the encounter with goblins\\ (a battle starts with players rolling for initiative which \\determines the order that they will take their turns)\end{tabular}                                          
\end{tabular}
} 
\caption{ \small Example dialogue transcript from D\&D game play with explanations.
}
\vspace{-0.4cm}
\label{tab:example_dialogues_explanations}
\end{table*}

\section{IDM Details}\label{appendix:IDM detail}
\paragraph{IDM Training}
We train two T5-3B models~\cite{raffel2020exploring} on our collected 2.5k human labeled dialogues, one for classifying DM texts that contain guidance or not (\emph{IDM-Identify}) and the other for extracting a sentence from the text (\emph{IDM-Extract}).
For \emph{IDM-Identify}, we treat the task as a binary prediction task and trains T5 to generate either 1 (contains guidance) or 0 (non-guidance) given the raw DM turn.
For \emph{IDM-Extract}, which is a harder task to select one sentence from the raw DM post as the most important guidance sentence, we have explored several approaches. 
We tried a text rewriting formulation that trains models to generate a special symbol (*) before and after a sentence in given text and an index selection formulation where we pass in DM turn indexed (\emph{e.g.}, ``1. A dwarf... 2. You notice some...'') and train the model to generate an index number (``2'').
Empirically we find the latter performs better.

\paragraph{IDM Model Evaluation}
We evaluate the IDM labeling performance on the test split of our human labels with 3-way redundancy on each label. 
We also tried other baselines for \emph{IDM-Extract}: 1) longest sentence; 2) last sentence; 3) 3-shot in-context learning using GPT-3 by asking them to select an index (same format as IDM); 4) encode each sentence and next-turn player action using SentenceBERT~\cite{reimers-2019-sentence-bert} and use cosine similarity to find the most similar sentence to the player action.
The \emph{IDM-identify} model reaches 82\% accuracy on binary classification tasks and \emph{IDM-extract} model reaches 70\% accuracy on a 10-way classification task (random guessing 10\%).
The best-performing baseline is 3-shot GPT-3 with in-context learning which reaches 55\%.
We argue that this task is hard and subjective as human agreements are very low.
However, experimental results on using IDM-generated labels (Table~\ref{tab:IDMAuto}) shows that it provides helpful signals and outperforms training on human labels significantly.
We also trained a DM model using GPT-3 labels and observe drops in performance overall.

\begin{figure}[tb]
	\centering
	\includegraphics[width=\columnwidth]{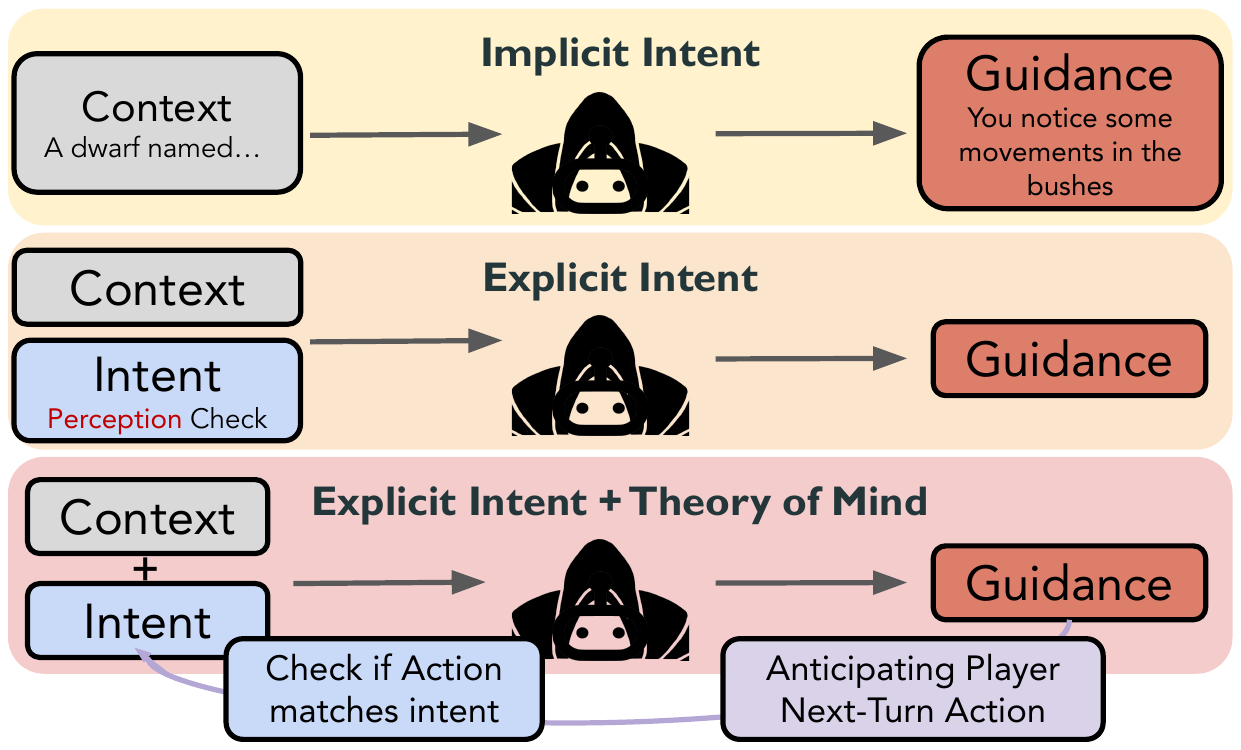}
	\caption{
	{Sketches of the three categories of methods}}
	\label{fig:sketch}
\end{figure}

\section{Human Guidance Annotation Details}\label{appendix:human label details}
\begin{figure*}[tb]
	\centering
	\includegraphics[width=1.0\linewidth]{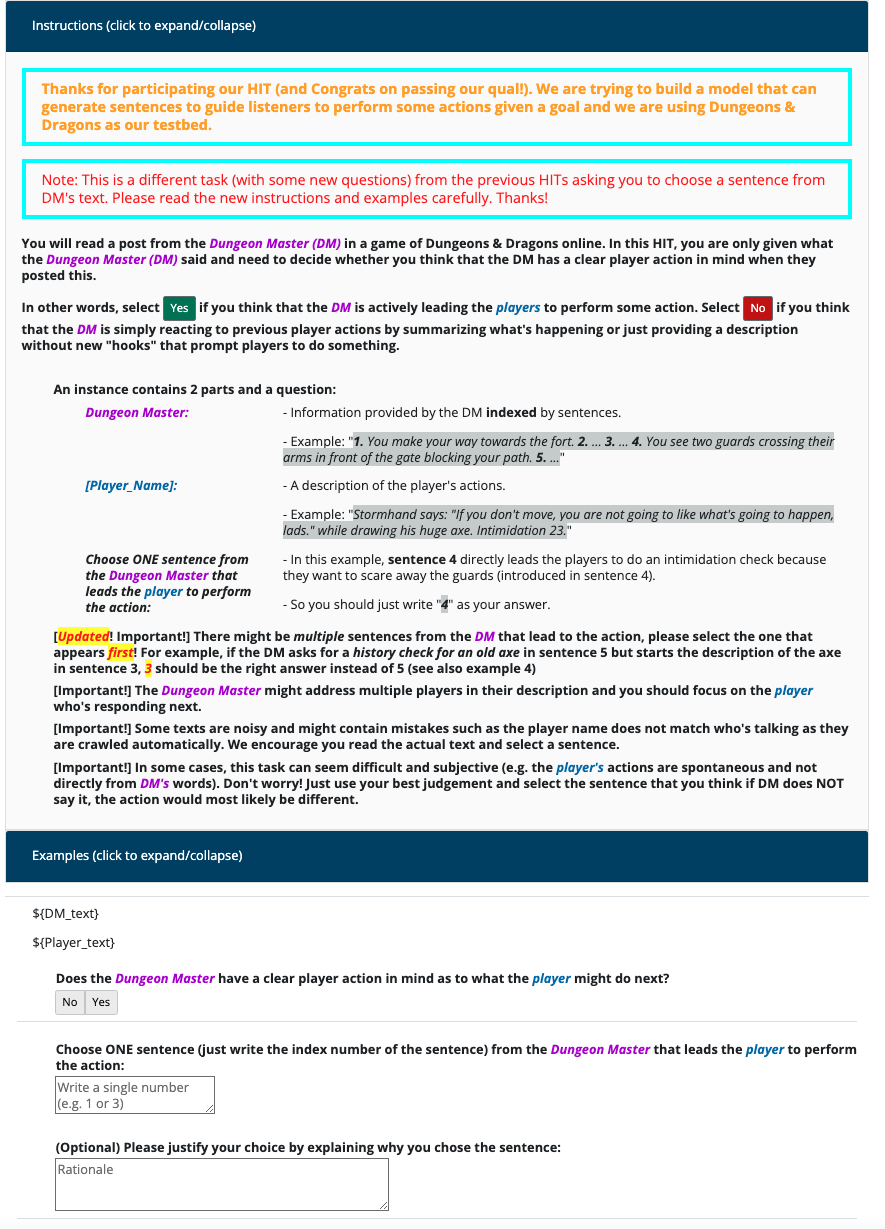}
	\caption{
	\textbf{Inference collection collecting guidance labels}.
	}
	\vspace{-0.1cm}
	\label{fig:idm1}
\end{figure*}

\begin{figure*}[tb]
	\centering
	\includegraphics[width=1.0\linewidth]{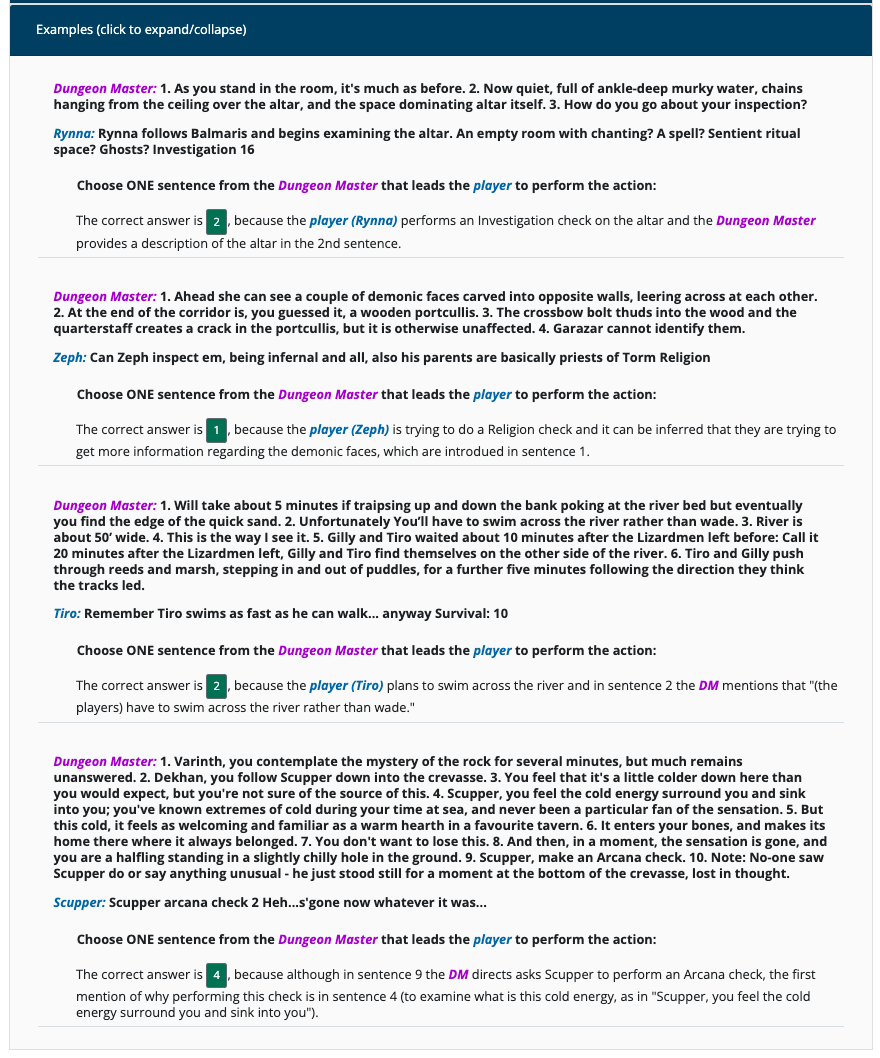}
	\caption{
	\textbf{Inference collection collecting guidance labels}.
	}
	\vspace{-0.1cm}
	\label{fig:idm2}
\end{figure*}

\begin{figure*}[tb]
	\centering
	\includegraphics[width=1.0\linewidth]{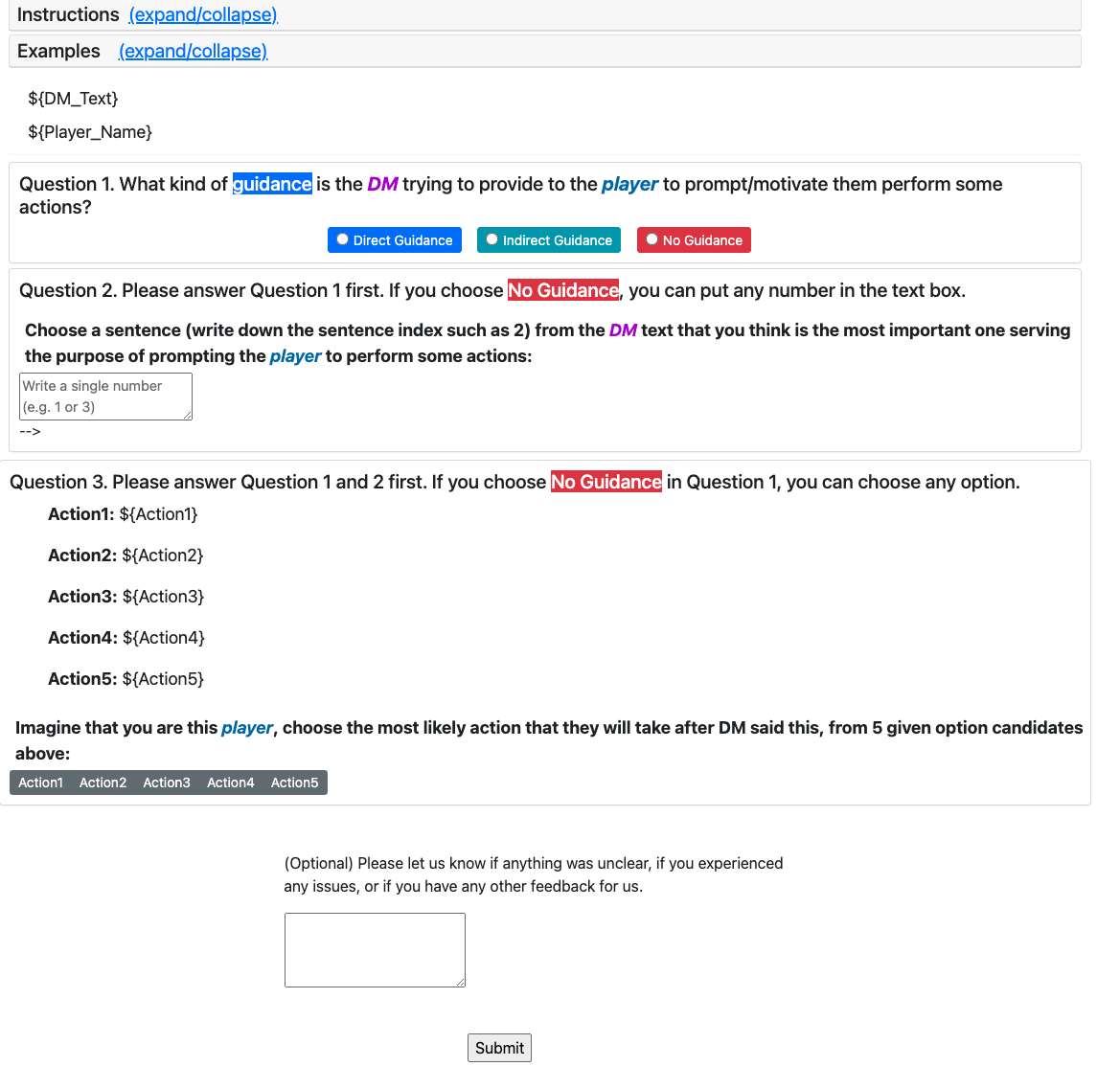}
	\caption{
	\textbf{Evaluation interface}.
	}
	\vspace{-0.1cm}
	\label{fig:eval}
\end{figure*}

Our designed human annotation interface for IDM labels and evaluation are included in Figures~\ref{fig:idm1},~\ref{fig:idm2}, and~\ref{fig:eval}.
We recruit around 120 AMT annotators from English-speaking countries (USC, UK, Australia, and New Zealand) since our data is in English. We first run a qualification test on 3 of our annotation questions and give qualifications to around 60 annotators who passed our test.
Then we provide detailed instructions and examples to them for completing our task. We also made it clear that our data is for research purposes and annotator ID will not be disclosed in any way.
Crowd working studies of standard NLP corpora (involving no personal disclosures) are not required by our IRB to be reviewed by them.

\section{Experimental and Model Details}
We train T5~\cite{raffel2020exploring} using Huggingface t5-trainer framework\footnote{\url{https://github.com/huggingface/transformers}}. All experimental results reported are a mean of 3 runs with different random seeds. We conduct a hyper-parameter search using a grid search for learning rates including 0.001, 0.0005, 0.0001, and 0.00005. We use a batch size of 4 for T5-3B and train on 2 NVIDIA RTX A6000 GPUs for around 30 hours or a batch size of 8 for T5-large (770M).

\section{Scientific Artifact Licensing}
The modeling framework~\cite{wolf2019huggingface,ramamurthy2022reinforcement}, and pre-trained models~\cite{raffel2020exploring} are open source.   The Play-By-Post dataset~\cite{callison-burch-tomar-et-al-2022} is used with permission of D\&D Beyond. We use these resources for non-commercial research purposes.



\begin{table*}[tb]
\centering
\small
\resizebox{\linewidth}{!}{
\begin{tabular}{c|l|l|l}
\textbf{Models}                                                                        & \multicolumn{1}{c|}{\textbf{Input Context}}                                                                                                                                                                                                                                               & \textbf{Input Intent (if any)}                                                                                                                                                                                                                                             & \multicolumn{1}{c}{\textbf{Output}}                                                                                                                                                                                              \\ \hline
Human-Label                                                                            & \multirow{8}{*}{\begin{tabular}[c]{@{}l@{}}\textbf{Kif (male halfling druid):} \\ \\ The frizzy hair helps her \\ blend into the background \\ and muffles the sound. \\ With everything looking \\ fairly lifeless do we see \\ any likely place to hole up?\\   \\ \textbf{Dungeon Master:}\end{tabular}} & N/A                                                                                                                                                                                                                                                                        & \begin{tabular}[c]{@{}l@{}}Kif, you've noticed a house on the north side\\  of the street you're on. It looks like it might \\ be occupied, and you've passed a few others.\end{tabular}                                         \\ \cline{1-1} \cline{3-4} 
IDM-Label                                                                              &                                                                                                                                                                                                                                                                                           & N/A                                                                                                                                                                                                                                                                        & \begin{tabular}[c]{@{}l@{}}You can talk to them; you don't have to. \\ You can act based on what you've already learned. \\ I'll wait for your response before continuing.\end{tabular}                                          \\ \cline{1-1} \cline{3-4} 
Mined Intent                                                                           &                                                                                                                                                                                                                                                                                           & \begin{tabular}[c]{@{}l@{}}The Dungeon Master intends to hint that \\ Kif should use his druidic abilities to \\ sense if there are any people or animals \\ inside the houses. If there are none, \\ then it's likely that the house is truly \\ unoccupied.\end{tabular} & \begin{tabular}[c]{@{}l@{}}Kif, you sense no people or animals in the \\ houses you've passed.\end{tabular}                                                                                                                      \\ \cline{1-1} \cline{3-4} 
Generated Intent                                                                       &                                                                                                                                                                                                                                                                                           & \begin{tabular}[c]{@{}l@{}}The Dungeon Master intends to guide the\\  player to investigate the houses that they \\ have already passed.\end{tabular}                                                                                                                      & \begin{tabular}[c]{@{}l@{}}You can go back and investigate the ones \\ you've already passed.\end{tabular}                                                                                                                       \\ \cline{1-1} \cline{3-4} 
ToM-RL-Mined                                                                           &                                                                                                                                                                                                                                                                                           & \begin{tabular}[c]{@{}l@{}}The Dungeon Master intends to hint that \\ Kif should use his druidic abilities to\\ sense if there are any people or animals \\ inside the houses. If there are none,\\ then it's likely that the house is truly \\ unoccupied.\end{tabular}   & \begin{tabular}[c]{@{}l@{}}Kif, you can sense if there are any people or \\ animals inside the houses. If there are none, \\ then it's likely that the house is truly unoccupied.\end{tabular}                                   \\ \cline{1-1} \cline{3-4} 
ToM-RL-Gen.                                                                            &                                                                                                                                                                                                                                                                                           & \begin{tabular}[c]{@{}l@{}}The Dungeon Master intends to guide the \\ player to investigate the houses that they \\ have already passed.\end{tabular}                                                                                                                      & \begin{tabular}[c]{@{}l@{}}You can investigate the houses you've already \\ passed, if you want to gain more information \\ about them.\end{tabular}                                                                             \\ \cline{1-1} \cline{3-4} 
\multicolumn{1}{l|}{\begin{tabular}[c]{@{}l@{}}Original DM \\ Transcript\end{tabular}} &                                                                                                                                                                                                                                                                                           & N/A                                                                                                                                                                                                                                                                        & \begin{tabular}[c]{@{}l@{}}Kif, you suspect that some of these houses, at least, are \\ genuinely unoccupied. You've only walked a couple of \\ blocks so far, and there may be a tavern somewhere in town.\end{tabular}        
\end{tabular}
} 
\caption{ Randomly sampled example outputs from different models.
}
\vspace{-0.3cm}
\label{tab:qulitative}
\end{table*}

\end{document}